\documentclass{article}
\usepackage{spconf,amsmath,graphicx,booktabs,pifont,multirow}
\usepackage{amsmath}
\usepackage{amssymb}
\usepackage{soul}
\usepackage[table]{xcolor}

\usepackage{threeparttable} 
\usepackage{subfigure}
\usepackage{makecell}
\usepackage{pifont}
\usepackage[accsupp]{axessibility}  
\usepackage[export]{adjustbox}
\usepackage{amsfonts} 
\usepackage[rightcaption]{sidecap}
\usepackage{xcolor}

\definecolor{americanrose}{rgb}{1.0, 0.01, 0.24}

\definecolor{americanrose}{rgb}{1.0, 0.01, 0.24}

\title{A Saliency Enhanced Feature Fusion based Multiscale RGB-D Salient Object Detection Network}
\name{Rui Huang$^{1}$ \quad Qingyi Zhao$^{1}$ \quad Yan Xing$^{2}$ \quad Sihua Gao$^{1}$\sthanks{Sihua Gao is the corresponding author with E-mail: shgao@cauc.edu.cn. This work was supported by the Fundamental Research Funds for Hebei Key Laboratory of Knowledge Computing for Energy \& Power under Grant HBKCEP202202 and Scientific Research Program of Tianjin Municipal Education Commission under Grant 2023KJ232.} \quad Weifeng Xu$^{3}$ \quad Yuxiang Zhang$^{1}$ \quad Wei Fan$^{1}$}

\address{$^{1}$ College of Computer Science and Technology, Civil Aviation University of China\\
$^{2}$ College of Safety Science and Engineering, Civil Aviation University of China\\
$^{3}$ Department of Computer, North China Electric Power University}

\begin{document}
%
\maketitle


\begin{abstract}
Multiscale convolutional neural network~(CNN) has demonstrated remarkable capabilities in solving various vision problems. However, fusing features of different scales always results in large model sizes, impeding the application of multiscale CNNs in RGB-D saliency detection. In this paper, we propose a customized feature fusion module, called \textit{Saliency Enhanced Feature Fusion} (SEFF), for RGB-D saliency detection. SEFF utilizes saliency maps of the neighboring scales to enhance the necessary features for fusing, resulting in more representative fused features.
Our multiscale RGB-D saliency detector uses SEFF and processes images with three different scales.
SEFF is used to fuse the features of RGB and depth images, as well as the features of decoders at different scales.
Extensive experiments on five benchmark datasets have demonstrated the superiority of our method over ten SOTA saliency detectors.
\end{abstract}

\section{Introduction}
\label{sec:intro}
RGB-D Salient object detection~(SOD) can use depth information to extract salient objects with similar colors to background in RGB images, which has wide applications in computer vision problems, such as object tracking~\cite{mahadevan2009saliency, ma2017saliency}, image retrieval~\cite{gordo2017end, zhang2016sketch}, and instance segmentation~\cite{wang2017saliency, sun2020mining}.

Many RGB-D SOD models aim to improve SOD performance by exploring effective multi-modal correlations. 
Early works focus on enhancing RGB features with depth maps.
Piao \textit{et al.}~\cite{piao2019depth} propose a depth refinement block using residual connections to fuse RGB and depth features.
Zhao \textit{et al.}~\cite{zhao2019contrast} enhance the contrast of the RGB features by multiplying them with an enhanced depth map.
Chen \textit{et al.}~\cite{chen2021rgb} propose pre-fusion across RGB and depth modalities, followed by in-depth feature fusion with 3D convolutions.
Now RGB-D SOD methods have proposed complex fusion architectures that refine RGB features and depth features simultaneously.
Wu \textit{et al}.~\cite{wu2023hidanet} suggest a multi-scale multi-level encoder fusion scheme with cross-domain supervision and decoder fusion, utilizing channel-wise dependencies.
Cong \textit{et al.}~\cite{cong2022cir} introduce a progressive attention-guided integration unit and importance-gated fusion, which integrates RGB and depth features in the encoder and decoder stages respectively.
Zhang \textit{et al.}~\cite{zhang2021bts} propose a bi-directional transfer-and-selection module to enable RGB and depth to mutually correct/refine each other in the encoder stage.
Although these complex fusion strategies improve RGB-D SOD performance, they also increase the size of models.

Recent studies have shown that multiscale Convolutional Neural Networks (CNNs) can achieve better performance in Super-resolution~\cite{lai2017deep, dong2016accelerating} and image deblurring~\cite{cho2021rethinking, kim2022mssnet} than single-scale CNNs. However, the use of multiscale CNNs in RGB-D saliency detection is hindered by the large model sizes and computations required. The main challenges of designing a multiscale CNN network for RGB-D SOD are: 1) \textbf{model size}. Although processing multiple scaled images, the designed multiscale network should have suitable model size and fast inference speed; 2) \textbf{information interchange across different scales}. Different scaled images can provide scale-specific features. How to efficiently exchange this valuable information across different scales is very challenging.

The main focus of our paper is on developing a module that can effectively fuse the features of RGB and depth images, as well as fuse features from different image scales. To achieve this, we propose \textit{Saliency Enhanced Feature Fusion} (SEFF) module, which utilizes saliency maps from neighboring scales to enhance the features required for fusion. This results in more representative fused features. Using SEFF, we have created a multiscale RGB-D saliency detector called SEFFsal. Our detector employs FasterNet~\cite{chen2023run} as the feature extraction backbone to extract features from both RGB and depth images. SEFF is then used to fuse these features, as well as the features of decoders from different scales.
\begin{figure*}[!tb]
	\centering
 	\includegraphics[width=0.96\linewidth]{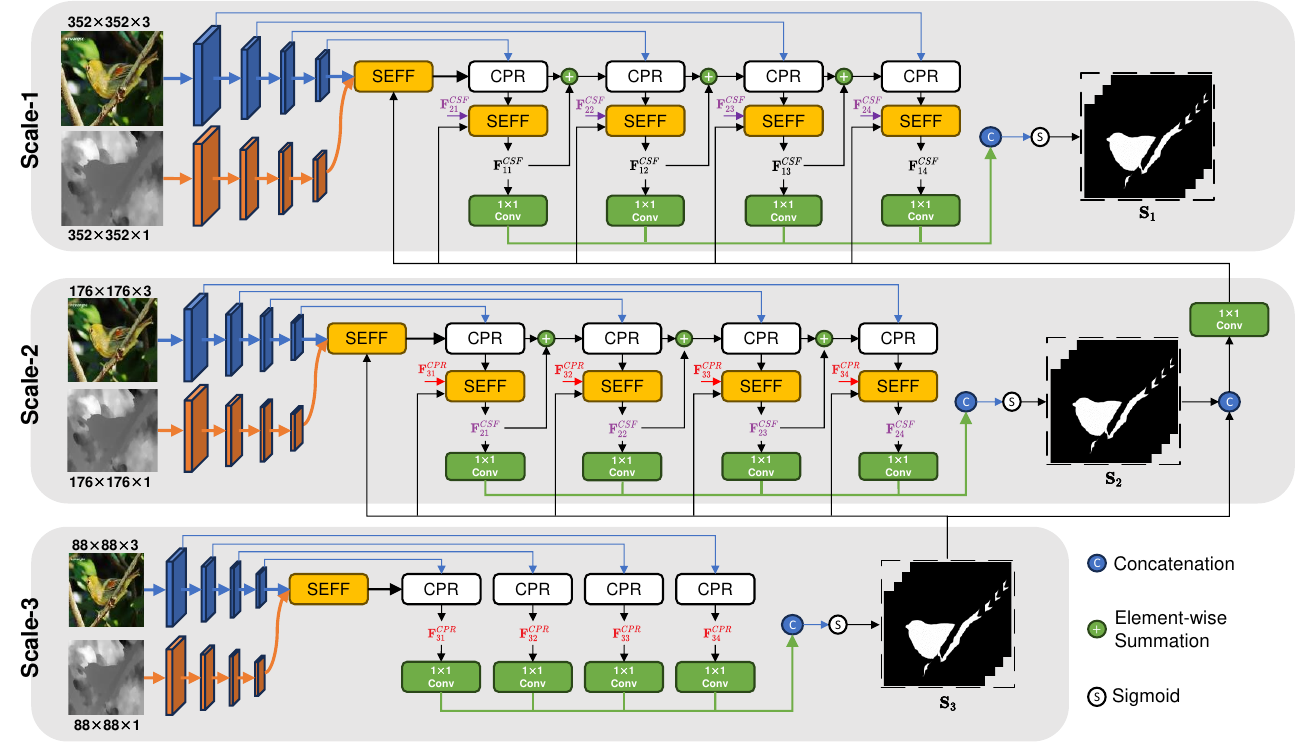}
	\caption{The framework of our multiscale RGB-D saliency detector.}
	\label{fig:framework}
\end{figure*}
We conducted numerous experiments on five benchmark datasets and compared our proposed SEFFsal with ten SOTA saliency detectors. The experimental results have demonstrated the effectiveness and efficiency of SEFFsal. 
The main contributions of this paper are as follows:
\begin{itemize}
\item We create an effective saliency enhanced feature fusion~(SEFF) module, which uses the saliency maps to improve the representative ability of the fused features.
\item Based on SEFF, we build a multiscale RGB-D saliency detector, which takes images with three different scales and generates high quality saliency results.
\item We have conducted extensive experiments on five benchmark datasets, which demonstrates that our method outperforms ten SOTA saliency detectors.
\end{itemize}

\section{Methodology}
\label{sec:propose}

\subsection{Overview}
We have developed a multiscale network called SEFFSal that uses an image $\mathbf{I}$ and its corresponding depth image $\mathbf{D}$ to detect salient objects. The overall architecture of SEFFSal is depicted in Fig.~\ref{fig:framework}. SEFFSal takes RGB and depth images at 3 different scales and employs FasterNet~\cite{chen2023run} as the fundamental feature extractor to extract features. We use image sizes of $352\times352$, $176\times176$, and $88\times88$ for the first, second, and third scales, respectively. To enable each feature extractor to adaptively extract features from the corresponding input, all feature extractors have independent parameters. Our \textit{Saliency Enhanced Feature Fusion}~(SEFF) module is responsible for fusing features of RGB and depth images and fusing the features of the decoders of different scales. In the following sections, we will explain the details of the SEFF module, our multiscale saliency detector, and its implementation.

\subsection{Saliency enhanced feature fusion module}
\begin{figure}[!tb]
	\centering
	 \includegraphics[width=\linewidth]{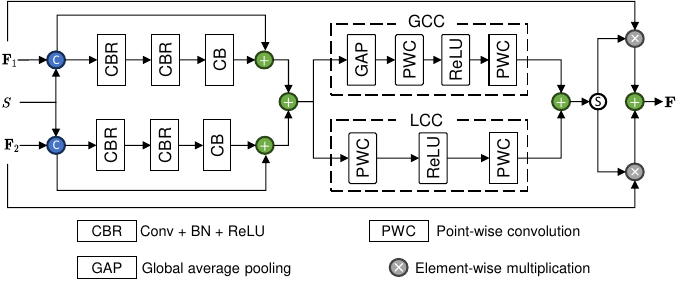}
	\caption{The detailed structure of the proposed SEFF.}
	\label{fig:csaf}
\end{figure}

The primary issue in a multiscale network is effectively combining features from various scales. To address this, we developed a feature fusion module called \textit{Saliency Enhanced Feature Fusion}~(SEFF). This module employs saliency maps to enhance the features and create more representative ones.
The detailed structure of SEFF can be observed in Fig.~\ref{fig:csaf}. Let $\mathrm{\mathbf{F}}_{1}$ and $\mathrm{\mathbf{F}}_{2}$ denote two features for fusion, and $\mathrm{\mathbf{S}}$ represents saliency maps. We first concatenate $\mathrm{\mathbf{S}}$ to $\mathrm{\mathbf{F}}_{1}$ and $\mathrm{\mathbf{F}}_{2}$, respectively, and refine them with several convolutional layers. These refined features are summarized together as the input of global and local channel context aggregators. The summation of local channel context~(LCC) and global channel context~(GCC) assigns importance to the features of $\mathrm{\mathbf{F}}_{1}$ and $\mathrm{\mathbf{F}}_{2}$ at both the channel-level and position-level. The final feature generated by SEFF is denoted as $\mathrm{\mathbf{F}}$.



The LCC aggregator adopts point-wise channel interactions for each spatial position, which only exploits point-wise channel interactions for each spatial position.  The GCC aggregator adopts global average pooling to achieve channel-level weights.
The whole process of SEFF is formalized as
\begin{equation}
    \mathrm{\mathbf{F}} = \Phi(\mathrm{\mathbf{F}}_{1}, \mathrm{\mathbf{F}}_{2}, \mathrm{\mathbf{S}}),
\end{equation}
where $\Phi(\cdot)$ denotes the fusion process of SEFF.

\subsection{SEFF-based multiscale RGB-D saliency detection}
With the proposed SEFF, we build our multiscale RGB-D saliency detection network. To improve the features, we have incorporated Compact Pyramid Refinement (CPR)~\cite{wu2021mobilesal} as the decoder module, which employs a 
multiscale learning strategy. We first use SEFF to integrate the features of RGB and depth images. Let $\mathrm{\mathbf{F}}_{ij}^{R}$ and $\mathrm{\mathbf{F}}_{ij}^{D}$ denote the RGB and depth feature of the $i$-th scale and $j$-th layer, and $\mathrm{\mathbf{S}}_{i}=Cat(\mathrm{\mathbf{S}}_{i1}, \mathrm{\mathbf{S}}_{i2}, \mathrm{\mathbf{S}}_{i3}, \mathrm{\mathbf{S}}_{i4})$ is the generated saliency maps of the $i$-th scale. The fused feature of the RGB and depth subnetwork can be generated as follows:
\begin{equation}
\begin{split}
    &\mathrm{\mathbf{F}}_{34}^{fusion} = \Phi(\mathrm{\mathbf{F}}_{34}^{R}, \mathrm{\mathbf{F}}_{34}^{D}, \mathrm{\mathbf{Z}}), \\
    &\mathrm{\mathbf{F}}_{24}^{fusion} = \Phi(\mathrm{\mathbf{F}}_{24}^{R}, \mathrm{\mathbf{F}}_{24}^{D}, \mathrm{\mathbf{S}}_{3}), \\
    &\mathrm{\mathbf{F}}_{14}^{fusion} = \Phi(\mathrm{\mathbf{F}}_{14}^{R}, \mathrm{\mathbf{F}}_{14}^{D}, Conv_1( Cat(\mathrm{\mathbf{S}}_{2}, \mathrm{\mathbf{S}}_{3}), 4)), \end{split}
\end{equation}
where $\mathrm{\mathbf{Z}}$ denotes zero maps, $Cat(\cdot)$ denotes concatenation operation, $Conv_1(\cdot,4)$ denotes $1\times1$ convolutional operation whose output channel number is 4. 
We experimentally found that fusing the RGB and depth features of the $4$-th layer can reduce the computation and has little affection on performance. Besides, we also found that better performance is achieved by concatenating the saliency maps of the $2$nd and $3$rd scale for the $1$st scale. Similarly, we use SEFF to fuse the decode features of adjacent scale as follows:
\begin{equation}
\begin{split}
    &\mathrm{\mathbf{F}}_{2j}^{CSF} = \Phi(\mathrm{\mathbf{F}}_{2j}^{CPR}, \mathrm{\mathbf{F}}_{3j}^{CPR}, \mathrm{\mathbf{S}}_{3}), \\
    &\mathrm{\mathbf{F}}_{1j}^{CSF} = \Phi(\mathrm{\mathbf{F}}_{1j}^{CPR}, \mathrm{\mathbf{F}}_{2j}^{CSF}, Conv_1(Cat(\mathrm{\mathbf{S}}_{2}, \mathrm{\mathbf{S}}_{3}), 4)), 
    \end{split}
\end{equation}
where the superscripts $CSF$ and $CPR$ denote the cross-scale fusion and features generated by the CPR layer, respectively. Please note that the fusion process is only carried out on the first and second scales.
\subsection{Implementation details}

\textbf{Saliency prediction.} We use multiple supervisions to efficiently supervise our network. Specifically, we generate the saliency maps from the features of the $3$rd scale $j$-th CPR module, $\mathrm{\mathbf{F}}_{3j}^{CPR}$, by a $1\times1$ convolutional layer as
\begin{equation}
    \mathrm{\mathbf{S}}_{3j} = \sigma( Conv_1(\mathrm{\mathbf{F}}_{3j}^{CPR}, 1)), 
\end{equation}
where $Conv_1(\cdot, 1)$ denotes $1\times1$ convolutional operation whose output channel number is 1. $\sigma(\cdot)$ denotes the sigmoid function. For the $1$st and $2$nd scales, we generate the saliency maps from features of SEFF modules by 
\begin{equation}
    \mathrm{\mathbf{S}}_{ij} = \sigma( Conv_1(\mathrm{\mathbf{F}}_{ij}^{CSF}, 1)), 
\end{equation}
where $i=1, 2$ and $j=1, 2, 3, 4$.


\noindent  \textbf{Loss.} In this paper, we use adaptive pixel intensity loss~\cite{lee2022tracer} to supervise the saliency prediction, which uses binary cross entropy, IoU, and L1 loss as
\begin{equation}
    \mathcal{L}_{API} = \lambda_1\mathcal{L}_{aBCE} + \lambda_2\mathcal{L}_{aIoU} + \lambda_3\mathcal{L}_{aL_1},
\end{equation}
where $\lambda_1=1$, $\lambda_2= 0.5$ and $\lambda_3 = 0.3$ in our experiments. The final loss function is defined as
\begin{equation}
     \mathcal{L} =\sum_{i=1}^{3}\sum_{j=1}^{4}\mathcal{L}_{API}(\mathrm{\mathbf{S}}_{ij},\mathrm{\mathbf{S}}_{GT}).
\end{equation}

\noindent \textbf{Super-parameters.} 
We utilize the Adam algorithm with a batch size of 10 and an initial learning rate of $5\times 10^{-5}$, which is decreased by a factor of 5 every 40 epochs, for a total of 100 epochs to optimize our network. During both the training and inference stages, we resize RGB and depth images to $352\times352$, $176\times176$, and $88\times88$ for saliency prediction without the use of any additional pre-processing or post-processing techniques. Our network is implemented using PyTorch and accelerated by a single NVIDIA 3090Ti GPU.

\begin{table*}[!htb]
	\addtolength{\tabcolsep}{-2pt}
	\setlength{\abovecaptionskip}{0.cm}
	\centering
	\caption{Quantitative comparison of different RGB-D SOD methods. The \textbf{bold} is the best. We use '-t', '-s', and '-m' to denote our method with tiny, small, and middle FasterNet backbones, respectively. '-scale1' and '-scale2' are two variants of our method.}
	\resizebox{\linewidth}{!}{
	\label{table:cmp_res}
	\begin{tabular}{rcccccccccccccccccccccccccc}  
		\toprule
            \multirow{2.5}{*}{Method} &  \multicolumn{4}{c}{LFSD} & \multicolumn{4}{c}{NJU2K} & \multicolumn{4}{c}{NLPR} & \multicolumn{4}{c}{SIP} & \multicolumn{4}{c}{STERE} & \multicolumn{4}{c}{AVG} & \multicolumn{1}{c}{Parameter} & \multicolumn{1}{c}{Speed}\\
            \cmidrule(lr){2-5}\cmidrule(lr){6-9}\cmidrule(lr){10-13}\cmidrule(lr){14-17}\cmidrule(lr){18-21}\cmidrule(lr){22-25}\cmidrule(lr){26-26}\cmidrule(lr){27-27}
            \multirow{2}{*}{ } & $ \mathcal{M\downarrow}$ & $F_\beta^\text{max}\uparrow$ & $E_\phi^\text{max}\uparrow$ & $S_\alpha\uparrow$ & $ \mathcal{M\downarrow}$ & $F_\beta^\text{max}\uparrow$ & $E_\phi^\text{max}\uparrow$ & $S_\alpha^\text{max}\uparrow$ & $ \mathcal{M\downarrow}$ & $F_\beta^\text{max}\uparrow$ & $E_\phi^\text{max}\uparrow$ & $S_\alpha^\text{max}\uparrow$ & $ \mathcal{M\downarrow}$ & $F_\beta^\text{max}\uparrow$ & $E_\phi^\text{max}\uparrow$ & $S_\alpha^\text{max}\uparrow$ & $ \mathcal{M\downarrow}$ & $F_\beta^\text{max}\uparrow$ & $E_\phi^\text{max}\uparrow$ & $S_\alpha^\text{max}\uparrow$ & $ \mathcal{M\downarrow}$ & $F_\beta^\text{max}\uparrow$ & $E_\phi^\text{max}\uparrow$ & $S_\alpha\uparrow$ & M & fps \\
            \midrule
            RD3D \cite{chen2021rgb}  & .134 & .703 & .780 & .739 & .035 & .918 & .953 & 0.921 & .033 & .874 & .936 & .899 & .094 & .768 & .851 & .779 & .050 & .875 & .924 & .886 & .069 & .827 & .889 & .845  & 46.90 & 54.62\\
            BBSNet \cite{zhai2021bifurcated}  & .122 & .752 & .820 & .766 & .022 & .963 & .980 & .954 & .023 & .922 & \textbf{.965} & .933 & .083 & .816 & .883 & .818 & .048 & .880 & .930 & .890 & .059  & .867  & .915  & .872 & 49.80 & 26.06\\
            MobileSal \cite{wu2021mobilesal} & .099 & .781 & .839 & .801 & .034 & .924 & .960 & .918 & .029 & .893 & .946 & .906 & .080 & .815 & .885 & .815 & .047 & .880 & .929 & .889 & .058 & .859 & .912 & .866 & 10.24 & 69.11\\
            BTSNet \cite{zhang2021bts}   & .098 & .803 & .855 & .824 & .023 & .961 & .980 & .952 & .029 & .898 & .952 & .919 & .057 & .878 & .919 & .872 & .049 & .888 & .935 & .896 & .051 & .886 & .928 & .892 & 100.17 & 23.22 \\
            SPNet \cite{zhou2021specificity}   & .118 & .772 & .843 & .771 & .016 & .965 & .981 & .958 & .020 & .923 & .964 & .930 & .096 & .772 & .866 & .782 & .043 & .892 & .939 & .896 & .059 & .865 & .919 & .867 & 175.29 & 12.42\\
            DCMF \cite{wang2022learning}  & .085  & .836  & .881  & .842  & .018  & .968  & .983  & .960  & .026  & .910  & .956  & .921 & .067  & .856  & .900  & .850  & .038  & .905  & .948  & .911  & .047 & .895 & .934 & .897 & 58.94 & 20.63\\
            SSLSOD \cite{zhao2022self}  & .083 & .821 & .868 & .838 & .026 & .962 & .969 & .950 & .032 & .862 & .906 & .900 & .085 & .822 & .862 & .837 & .047 & .884 & .917 & .897 & .055 & .870 & .904 & .884  & 74.17 & 52.41 \\
            CIRNet \cite{cong2022cir}  & .118 & .789 & .845 & .785 & .029 & .953 & .976 & .945 & .024 & .920 & .961 & .930 & .086 & .820 & .886 & .824 & .053 & .886 & .933 & .887 & .062 & .873 & .920 & .874 & 103.15 & 30.91 \\ 
            HiDANet \cite{wu2023hidanet}  & .121  & .760  & .829  & .771 & .018  & .962  & .980  & .952 & .021  & .925  & .964  & .931 & .093  & .793  & .883  & .788 & .046  & .885  & .933  & .889 & .060 & .865 & .918 & .866 & 130.64 & 9.42\\
            PopNet \cite{wu2022source}  & .079 & .828 & .875 & .844 & \textbf{.014} & .970 & .984 & .961 & .020 & .921 & .964 & .928 & .051 & .888 & .924 & .878 & .033 & .914 & .951 & .916 & .040 & .904 & .939 & .905 & 223.88 & 10.06 \\
            {Ours-m} & \textbf{.062}  & \textbf{.869}  & \textbf{.906} & \textbf{.870} & \textbf{.014}  & \textbf{.974}  & \textbf{.986} & \textbf{.965} & \textbf{.019}  &  \textbf{.927}  & \textbf{.965} & \textbf{.937}  & \textbf{.047}  & \textbf{.898}  & \textbf{.926} & \textbf{.885} & \textbf{.032}  & \textbf{.917}  & \textbf{.953} & \textbf{.921} & \textbf{.035} & \textbf{.917} & \textbf{.947} & \textbf{.915} & 498.94 & 12.41\\
            \cmidrule(r){1-27}
             \rowcolor{gray!30}{Ours-s} & .064  & .871  & .906  & .865 & .016  & .966  & .984  & .959 & .020  & .923  & .963  & .934 & .061  & .862  & .900  & .849 & \textbf{.032}  & .914  & \textbf{.953}  & .917 & .039 & .907 & .941 & .905 & 325.35 & 14.80 \\
              \rowcolor{gray!30}{Ours-t} & .080 & .836 & .886 & .837 & .022 & .953 & .971 & .943 & .023 & .918 & .960 & .924  & .078 & .819 & .868 & .810 & .038 & .904 & .942 & .902 & .048 & .886 & .925 & .883 & 163.27 &  16.56\\
            \cmidrule(r){1-27}
               \rowcolor{gray!10}{Ours-scale1} & .085 & .819 & .861 & .828 & .016 & .968 & .984 & .959 & .020 & .919 & .960 & .930 & .049 & .893 & .922 & .878 & .035 & .910 & .945 & .914 & .041 & .902 & .934 & .902 & 145.93 & 42.04\\
                \rowcolor{gray!10}{Ours-scale2} & .071  & .854  & .892  & .857  & .015  & .971  & .985  & .963  & .020  & .926  & .964  & .935  & .050  & .889  & .920  & .877  & .033  & .916  & .951  & .919  & .038  & .911  & .942  & .910 & 322.44 & 18.79 \\
            \cmidrule(r){1-27}
                \rowcolor{gray!30}{w/o SEFF}  & .091 & .844 & .875 & .846 & .039 & .946 & .972 & .929 & .034 & .896 & .952 & .910 & .077 & .846 & .897 & .836 & .054 & .893 & .941 & .894 & .059 & .885 & .927 & .883  & - & - \\
		\bottomrule
	\end{tabular}
        }\vspace{-10px}
\end{table*}


\section{Experiment}
\subsection{Setup}
\textbf{Baseline.} Ten SOTA RGB-D saliency detectors are used for comparison, including RD3D \cite{chen2021rgb},   BBSNet \cite{zhai2021bifurcated}, MobileSal \cite{wu2021mobilesal},  BTSNet \cite{zhang2021bts},  SPNet \cite{zhou2021specificity},  DCMF \cite{wang2022learning},  SSLSOD \cite{zhao2022self},  CIRNet \cite{cong2022cir}, HiDANet \cite{wu2023hidanet}, and PopNet \cite{wu2022source}. 
Each of these methods was retrained using their default setting, except for image size ($352\times352$) and number of epochs ($100$).


\textbf{Datasets.} We conducted all experiments on 5 benchmark datasets, i.e., LFSD \cite{li2014saliency},
NJU2K \cite{ju2014depth}, NLPR \cite{peng2014rgbd},  SIP \cite{fan2020rethinking} and STERE \cite{niu2012leveraging}. The partition of the training and testing datasets is the same as that proposed in \cite{zhao2019contrast, zhai2021bifurcated}. 

\textbf{Evaluation metrics.} 
We use Mean Absolute Error~($\mathcal{M}$), max F-measure~($F_\beta^\text{max}$), max E-measure~($E_\phi^\text{max}$), and S-measure~($S_\alpha$) to quantitatively evaluate the performance of our method. We set $\beta$ to 0.3, $\alpha$ to 0.5, and $\phi$ refers to the enhanced-alignment matrix as presented in \cite{fan2018enhanced}.

\subsection{Results and Analysis}
\begin{figure}[!tb]
	\centering
	 \includegraphics[width=\linewidth]{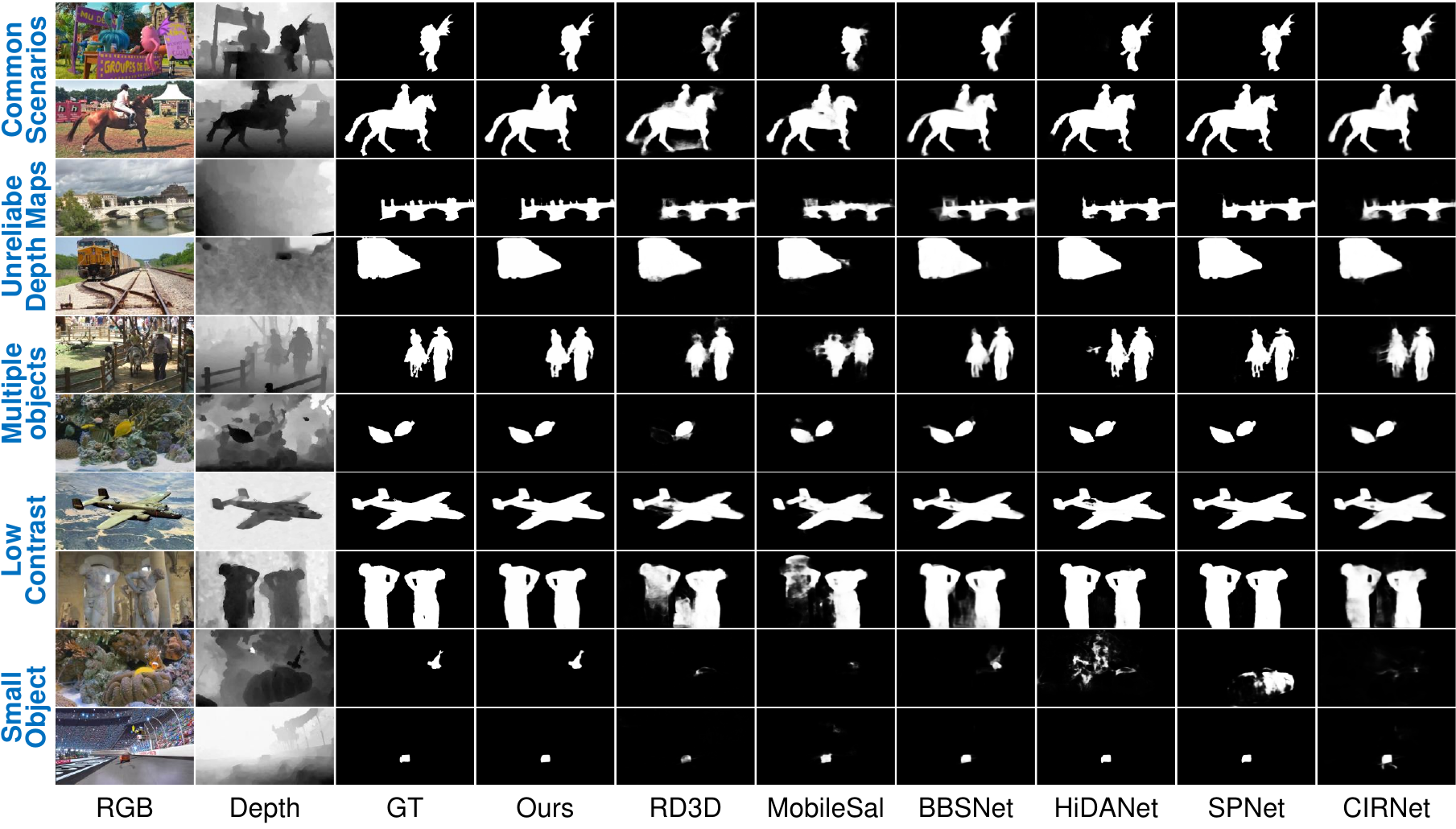}\vspace{-5px}
 \caption{Some typical results of different RGB-D SOD methods on various scenes.}
	\label{fig:res}\vspace{-10px}
\end{figure}

Fig.~\ref{fig:res} shows some typical saliency detection results on five scenes, such as common scenarios, unreliable depth maps, multiple objects, low contrast, and small objects. Compared with the existing saliency detectors, our method can produce more complete salient objects (e.g., the $8$-th and $9$-th rows) and clearer backgrounds (e.g., the $2$-nd and $5$-th rows).
Table.~\ref{table:cmp_res} shows the quantitative results of the different RGB-D SOD methods on five datasets.
We can see that all values are better than the values of the compared methods. Among the compared methods, PopNet is ranked in the first place.
%
%
Compared with PopNet across the five datasets, our method~(i.e., Ours-m) achieves 10.81\%, 1.55\%, 0.64\%, and 1.1\% relative improvements in terms of Mean Absolute Error, max F-measure, max E-measure, and S-measure, respectively. Both the qualitative and quantitative results demonstrate the effectiveness and superiority of our method.


\subsection{Ablation study}

\textbf{The effectiveness of SEFF.}
We tested the effectiveness of SEFF by replacing it with several CBR blocks, including convolution, batch normalization, and ReLU. These convolutional blocks have the same number of parameters as SEFF. The results shown in Table.~\ref{table:cmp_res}, indicates that without SEFF, the performance of our method dropped significantly. This demonstrates the effectiveness of SEFF.



\textbf{The performance of SEFFsal with different scales.} 
We analyzed two versions of our approach, namely \textit{scale1} which solely employs the network structure of the third scale, and \textit{scale2} which utilizes the network architecture of the second and third scales. It's worth noting that the maximum input size is $352\times352$. Based on the results presented in Table.~\ref{table:cmp_res}, we find that incorporating multiple scales leads to improved SOD performance. In fact, the network comprising three scales outperformed those with only one or two scales.



%

\textbf{The performance of SEFFsal with different base models.}
Our method, SEFFsal, utilizes FasterNet~\cite{chen2023run} as the primary feature extractor. Table.~\ref{table:cmp_res} reports the results of SEFFsal with different base models. It is evident from the table that our approach outperforms most of the compared methods, even with the use of a small model, as seen in Ours-t. Our method, Ours-s, has already ranked first among all the compared methods. Additionally, using a larger model like Ours-m can further enhance the detection performance.


\textbf{Parameter and speed.}
According to Table.~\ref{table:cmp_res}, we find that our multiscale model has more parameters than single-scale models. However, our method has a faster inference speed than both HiDAnet and PopNet single-scale models.






\section{Conclusion}
In this paper, we proposed a multiscale RGB-D salient object detection network based on a novel and effective feature fusion module, \textit{Saliency Enhanced Feature Fusion} (SEFF). This module uses saliency maps to enhance the features required for fusion, resulting in more representative fused features.  We utilize SEFF to fuse the features of RGB and depth images, as well as the features of decoders at different scales.
Through extensive experiments on five benchmark datasets, we have demonstrated that our method outperforms ten state-of-the-art saliency detectors. We plan to explore a lightweight multiscale network for RGB-D SOD in future work.

\vfill
\newpage


\small
\bibliographystyle{IEEEbib}
\bibliography{references}

\end{document}